\pdfoutput=1

\documentclass[11pt]{article}

\usepackage{emnlp2021}

\usepackage{times}
\usepackage{latexsym}

\usepackage[T1]{fontenc}
\usepackage[utf8]{inputenc}

\usepackage{microtype}

\usepackage{todonotes}
\usepackage{booktabs}
\usepackage{cleveref}
\crefformat{footnote}{#2\footnotemark[#1]#3}
\usepackage{enumitem}
\usepackage{float}
\usepackage{multirow}
\usepackage{subfig}
\usepackage[export]{adjustbox}
\usepackage{colortbl}
\newcommand{\stdintable}[1] {\textcolor{gray}{\scriptsize{$\pm$#1}}}

\title{AdapterDrop: On the Efficiency of Adapters in Transformers}

\author{Andreas R\"uckl\'e\thanks{~~Work done prior to joining Amazon.} \and Gregor Geigle \and Max Glockner, \\
\textbf{Tilman Beck \and Jonas Pfeiffer \and Nils Reimers \and Iryna Gurevych}
\\
	Ubiquitous Knowledge Processing Lab (UKP)\\
	Department of Computer Science, Technische Universit\"{a}t Darmstadt\\
	{\url{www.ukp.tu-darmstadt.de}}
} 

\date{}

\begin{document}
\maketitle
\begin{abstract}
Transformer models are 
expensive to fine-tune, slow for inference, and have large storage requirements. Recent approaches tackle these shortcomings by training smaller models, dynamically reducing the model size, and by training light-weight adapters. In this paper, we propose \textbf{AdapterDrop}, removing adapters from lower transformer layers during training and inference, which incorporates concepts from all three directions. %
We show that %
AdapterDrop
can dynamically reduce the computational overhead when performing inference over multiple tasks simultaneously, with %
minimal
decrease in task performances. We further prune adapters from AdapterFusion, which improves the inference efficiency while maintaining the task performances entirely. 
\end{abstract}

\section{Introduction}

While transfer learning has become the go-to method for solving NLP tasks \cite{pan2009survey, torrey2010transfer, ruder2019neural, howard18ulmfit, peters18elmo}, transformer-based models are notoriously deep requiring millions or even billions of parameters  \cite{radford2018gpt, devlin-bert, radford2019gpt3, liu2019roberta, brown20gpt3}.
This results in slow inference 
and 
large storage
 requirements.

At least three independent lines of research have recently evolved to tackle these shortcomings.
(1)~Smaller and faster models that are either distilled or trained from scratch \cite{Sanh2019Distilbert, Sun2020mobilebert,bai2020binarybert,wang2020minilm}. (2)~Robustly trained transformers in which the model depth can be reduced at run-time, thereby decreasing inference time dynamically~\cite{Fan2020LayerDrop,elbayad2020depth,JiACL2020,houNeurIPS2020}. (3)~Adapters, which, instead of fully fine-tuning the model, only train a newly introduced set of weights at every layer, thereby sharing the majority of parameters between tasks~\cite{Houlsby2019adapters,bapna-firat-2019-simple,pfeiffer2020AdapterHub}.
Adapters have been shown to work well for machine translation~\citep{bapna-firat-2019-simple}, cross-lingual transfer~\cite{pfeiffer20madx,Pfeiffer2020unks, ustun2020udapter, Vidoni2020OrthogonalLA, Ansell2021MADG}, community QA~\citep{ruckle-etal-2020-multicqa}, and task composition for transfer learning~\cite{Stickland019bertpals, Pfeiffer2020adapterfusion,Lauscher2020comonsense, wangk2020adapters,poth2021pre}.
Despite their recent popularity, the computational efficiency of adapters has not been explored beyond parameter efficiency. 
We close this gap and establish the computational efficiency of two adapter architectures at training and inference time. 
We investigate different strategies to further improve the efficiency of adapter-based models by incorporating ideas from all three directions mentioned above. %
Our strategies rely on dropping out adapters from transformers, at training and inference time, 
resulting
in models that are dynamically adjustable regarding the available computational resources. 
Our approaches are agnostic to the pre-trained transformer model (e.g., base, large), which makes them broadly applicable. 

\paragraph{Contributions:}
\begin{enumerate}[leftmargin=*,noitemsep,partopsep=0pt,topsep=0pt,parsep=0pt]
    \item We are the first to establish the computational efficiency of adapters compared to full fine-tuning. We show that the training steps of adapters can be up to $60\% $ 
    faster than full model fine-tuning with common hyperparameter choices, while being 4--6\% slower at inference.
    Hence, adapters are a suitable choice for researchers interested in achieving faster training times, or when requiring extensive hyperparameter tuning. 
    
    \item We propose AdapterDrop, the efficient and dynamic removal of adapters with minimal impact on the task performances. We show that dropping adapters from lower transformer layers considerably improves the inference speed in multi-task settings. For example, with adapters dropped from the first five layers, AdapterDrop is %
    $39\%$ faster when performing inference on 8 tasks \emph{simultaneously}.
    This can be beneficial for researchers working on models that need to make multiple predictions on each input. 
    \item We prune adapters from adapter compositions in AdapterFusion~\citep{Pfeiffer2020adapterfusion} and retain only the most important adapters after transfer learning,
    resulting in faster inference while maintaining the task performances entirely.
    This is suitable for settings with little labeled training data, where AdapterFusion can achieve ample improvements over standard single task models.
\end{enumerate}

\section{Efficiency of Adapters}

We first establish the computational efficiency of adapters \emph{without} AdapterDrop. As illustrated in Figure~\ref{fig:AdapterDrop}, significant differences exist in the forward and backward pass when fine-tuning adapters compared to fully fine-tuning the model. In the forward pass, adapters add complexity with the additional components; however, it is not necessary to backpropagate through the entire model during the backward pass.
We compare the training and inference speed of full model fine-tuning against the adapter architectures of \citet{Houlsby2019adapters}  and \citet{Pfeiffer2020adapterfusion}
(depicted in Figure~\ref{fig:AdapterDrop}) 
using
the  \href{https://AdapterHub.ml}{AdapterHub.ml} framework%
~\cite{pfeiffer2020AdapterHub}. %
We conduct our measurements 
with the transformer configuration of BERT base and verify them with different GPUs.\footnote{We experiment with newer and older GPUs, Nvidia V100 and Titan X, respectively. See Appendix~\ref{sec:appendix:measurements} for details.}

We provide measurements corresponding to common experiment configurations in Table~\ref{tab:adapters:train_inference_speedup}.

\begin{table}
\footnotesize
    \centering
    \setlength{\tabcolsep}{5pt}
\begin{tabular}{@{}ll*{4}{r}@{}}
	\toprule
	\bf Setting & \bf Adapter &	\multicolumn{4}{c}{\textbf{Relative speed} \textit{(for Seq.Len./Batch)}} \\
	\cmidrule(l){3-6}
				&			& 128/16	& 128/32	& 512/16	& 512/32 \\
	\midrule
	Training	& Houlsby	& 1.48		& 1.53		& 1.36	& 1.33 \\
				& Pfeiffer	& 1.57		& 1.60		& 1.41	& 1.37 \\
	Inference	& Houlsby	& 0.94		& 0.94		& 0.96 & 0.96 \\
				& Pfeiffer	&0.95		& 0.95		& 0.96	& 0.96 \\
	\bottomrule
\end{tabular}
    \caption{Relative speed of adapters compared to fully fine-tuned models. For example, 1.6 for training with the Pfeiffer adapter means that we can perform 1.6 training steps with this adapter in the time of one training step with full model fine-tuning.
    }
    \label{tab:adapters:train_inference_speedup}
\end{table}

\paragraph{Training.}

Adapters can be considerably faster compared to full model fine-tuning---60\% faster in some configurations. 
The two adapter architectures differ only marginally in terms of training efficiency: due to its simpler architecture, training steps of the Pfeiffer adapters are slightly faster. %
The magnitude of the differences depends on the input size; the available CUDA cores are the primary bottleneck.\footnote{We include detailed plots in Appendix~\ref{sec:appendix:efficiency-adapters}.} %
We do not observe any particular differences between adapters and full fine-tuning regarding the training convergence.\footnote{We also pre-train adapters with masked language modeling, finding that this does not yield better  results (Appendix~\ref{app:ssec:adapter_init}).}

The training speedup can be explained by the decreased overhead of gradient computation. 
Most of the parameters are frozen when using adapters
and it is not necessary to backpropagate through the first components
(see Figure~\ref{fig:AdapterDrop}).

\paragraph{Inference.}
The two adapter architectures %
are  94--96\% as fast as 
fully fine-tuned models, which varies depending on the input size. %
This can have a considerable impact when deployed at scale.

\section{AdapterDrop}
\label{sec:adapter-drop}
We have established that adapters are more efficient in terms of training time, however, there is a perpetuate need for sustainable and efficient models~\cite{StrubellGM19}. 
Backpropagating through as few layers as possible would further improve the efficiency of \emph{training} adapters. %
The efficiency for \emph{inference} can be improved by %
sharing representations at lower transformer layers %
when simultaneously performing inference for multiple tasks---in other words, when performing multiple independent classifications on the same input. 
We establish this in Table \ref{tab:adapterdrop-speedup}, finding that models are up to $8.4\%$ faster %
with \textbf{every} shared layer (16 tasks). 

\begin{figure}[t!]
    \centering
         \centering
        \includegraphics[width=0.95\linewidth]{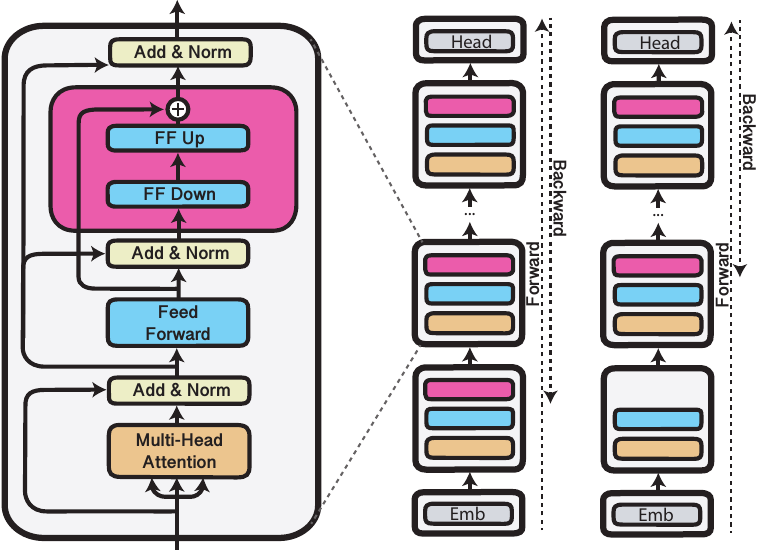}
        \caption{Standard adapter fine-tuning vs. AdapterDrop fine-tuning. The left model includes adapters at every layer whereas the right model has adapters dropped at the first layer. The arrows to the right of each model indicate the information flow for the \textit{Forward} and \textit{Backward} pass through the model. }
        \label{fig:AdapterDrop}
\end{figure}

Motivated by these observations, we propose AdapterDrop: Dynamically removing adapters from lower transformer layers (depicted in  Figure~\ref{fig:AdapterDrop}).
AdapterDrop is similar to dropping out entire transformer layers~\citep{Fan2020LayerDrop}, however, specialized to adapter settings---%
where lower layers often have a small impact on the task performances~\citep{Houlsby2019adapters}.

\begin{table}
\footnotesize
    \centering
\begin{tabular}{lrrrr}

\toprule

\bf Simultaneous Tasks & 2 & 4 & 8 & 16\\
\midrule
{\bf Speedup} (each layer) & 4.3\%& 6.6\% & 7.8\% & 8.4\% \\

\bottomrule
\end{tabular}
    \caption{Speedup for each shared transformer layer when performing inference for multiple tasks simultaneously (details are given in Appendix~\ref{sec:appendix:adapterdrop})
    }
    \label{tab:adapterdrop-speedup}
\end{table}

\begin{figure}
	\centering
	\includegraphics[width=0.95\columnwidth]{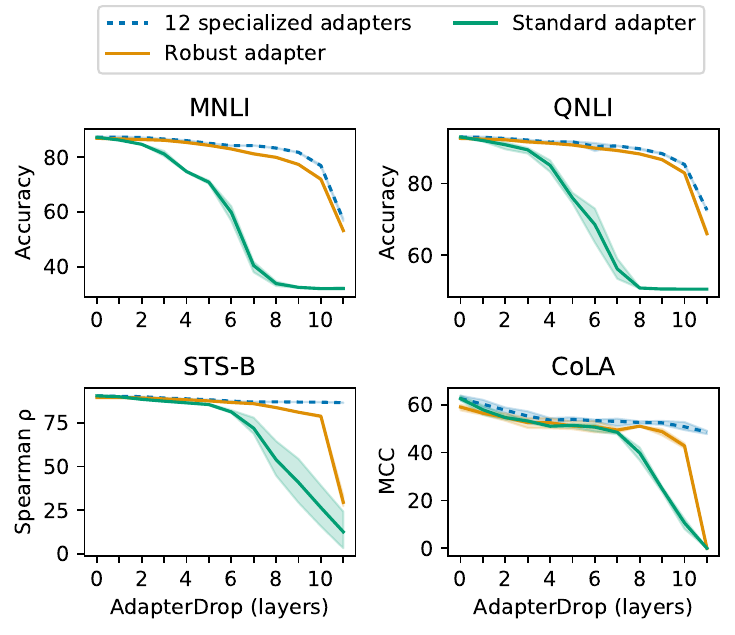}
	\caption{
	Task performances in relation to dropped 
	layers during evaluation (Figure~\ref{fig:adapter-drop-large} shows all tasks). `Standard adapter' %
	is trained with no dropped layers.
    }
	\label{fig:adapter-layer-drop}
\end{figure}

We study two training methods for AdapterDrop: (1)~\textbf{Specialized} AdapterDrop: Removing adapters from the first $n$ transformer layers, where $n$ is fixed during training. This yields separate models \emph{for each} possible $n$.
(2)~\textbf{Robust} AdapterDrop: Drawing the integer $n$ randomly from $[0, 11]$ for each training batch.\footnote{We also explored dropping adapters from randomly chosen layers (instead of early layers). This generally performs worse and it requires selecting a suitable dropout rate.} This yields \emph{one robust} model that is applicable to a varying number of dropped layers.
We study the effectiveness of AdapterDrop on the devsets of the GLUE benchmark~\cite{wang18glue} using RoBERTa base~\cite{liu2019roberta}.\footnote{The detailed setup is listed in Appendix~\ref{sec:appendix:task-performances}.}

Figure~\ref{fig:adapter-layer-drop} shows that specialized AdapterDrop maintains good results even with several dropped layers. 
With the first five layers dropped, specialized AdapterDrop maintains 97.1\% of the original performance %
(averaged over all eight GLUE tasks; see Table~\ref{tbl:adapter-drop-maintained-performance}). %
Moreover,  robust AdapterDrop  achieves comparable results, 
and with five  layers dropped it maintains 95.4\% of the original performance (on avg).
The  advantage of \textit{robust} over \textit{specialized} AdapterDrop is that the \textit{robust} variant can be dynamically scaled. %
Based
on current available computational resources, \textit{robust} AdapterDrop can (de)activate layers with the same set of parameters, whereas \textit{specialized} AdapterDrop needs to be trained for every setting explicitly.

The efficiency gains can be large. When performing inference for multiple tasks simultaneously, we measure inference speedups of 21--42\% with five dropped layers---depending on the number of simultaneous tasks (Table~\ref{tab:adapterdrop-speedup}).\footnote{For more details see Appendix~\ref{sec:appendix:adapterdrop}}
\emph{Training} of our robust adapters is also more efficient, which increases the speed of training steps by 26\%.\footnote{Every dropped adapter improves the speed of training steps by 4.7\% and we drop on average 5.5 adapters when training robust adapter models (more hyperparameter settings and details are given in Appendix~\ref{sec:appendix:adapterdrop}).}

\section{Efficiency of AdapterFusion}
\label{sec:adapter-fusion}
\begin{figure}[!t]
         \centering
        \includegraphics[width=0.95\linewidth]{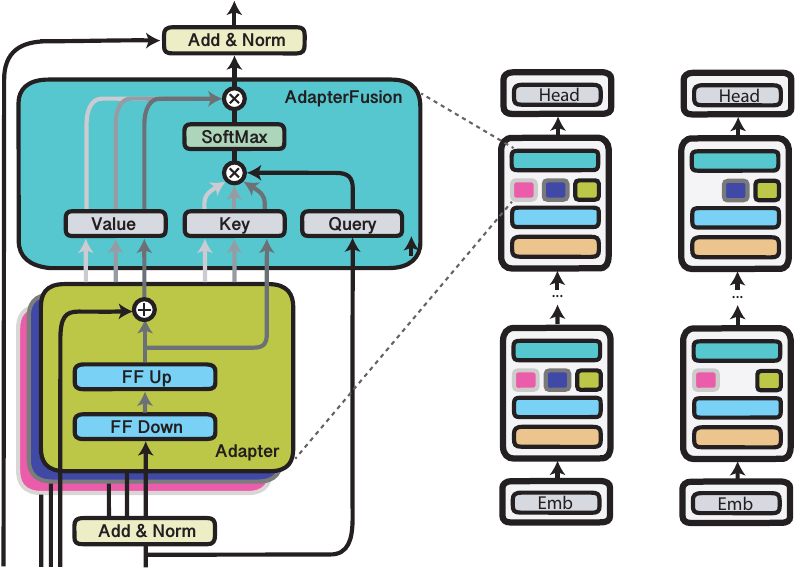}
        \caption{Standard AdapterFusion vs. AdapterFusion pruning, each with 3 adapters initially. The left model includes all adapters at every layer whereas the right model has one adapter pruned at every layer. %
        }
        \label{fig:AdapterFusionDrop}
\end{figure}

AdapterFusion~\citep{Pfeiffer2020adapterfusion} leverages the knowledge of \emph{several} adapters from different tasks and learns an optimal combination of the adapters' output representations for a single target task (see Figure~\ref{fig:AdapterFusionDrop}). AdapterFusion~(AF) is particularly useful for small training sets where learning adequate models is difficult. %
Despite its effectiveness, AF is computationally expensive because all included adapters are passed through sequentially.\footnote{We also test AF with parallel operations and found no efficiency gains (see Appendix \ref{sec:appendix:parallel-fusion}).}

Table~\ref{tab:fusion-speed-adapters} shows that the differences can be substantial for both training and inference. For instance, compared to a fully fine-tuned model, AF with eight adapters is around 47\% slower at training time and 62\% slower at inference.\footnote{All with Pfeiffer adapter and depending on the input size. We provide more measurements in Appendix~\ref{sec:appendix:efficiency-adapterfusion}.}

\begin{table}
\footnotesize
    \centering
    \setlength{\tabcolsep}{5pt}
    \begin{tabular}{rrrrr}
\toprule
  & \multicolumn{2}{c}{\bf AF vs. Full FT} & \multicolumn{2}{c}{\bf AF vs. Adapter}  \\
  \cmidrule(lr){2-3} \cmidrule(l){4-5}
   \bf Adapters &  Training & Inference & Training & Inference \\
\midrule
\bf 2 & 0.92 & 0.64 & 0.57 & 0.68\\
\bf 8 & 0.53 & 0.38 & 0.33 & 0.40\\
\bf 16 & 0.33 & 0.24 & 0.21 & 0.26\\
\bottomrule

\end{tabular}

    \caption{Relative speed of AdapterFusion (with 2/8/16 adapters) compared to a fully fine-tuned model and compared to a single-task adapter (right). Measured with a batch size of 32, and a sequence length of 128. %
    }
    \label{tab:fusion-speed-adapters}
    
\end{table}

\section{AdapterDrop for AdapterFusion}
There exists considerable potential for improving the efficiency of AF, especially at \emph{inference} time.
We address this with two variants of AdapterDrop for AF by (1)~removing entire AF layers; (2)~pruning the least important adapters from AF models.

\subsection{Removing AdapterFusion Layers}
We fuse the adapters from all eight GLUE tasks and observe the largest gains of AF %
on RTE and CoLA. 
We additionally train robust AF models with the same procedure as in \S\ref{sec:adapter-drop}.
We investigate from how many lower layers we can remove AF at test time while still outperforming the corresponding single-task adapter (without AdapterDrop).
\begin{figure}
	\centering
	\includegraphics[width=0.91\columnwidth]{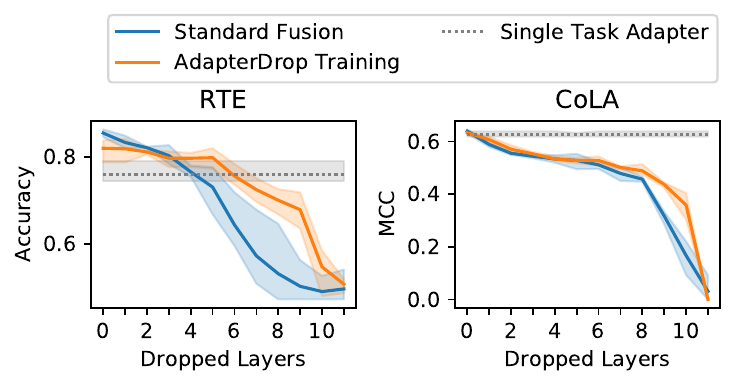}
	\caption{Comparison of AdapterFusion with (orange) and without (blue) AdapterDrop training during inference when omitting early AF layers.}
	\label{fig:fusion-early-layer-dropout-cola-rte}
\end{figure}

Figure~\ref{fig:fusion-early-layer-dropout-cola-rte} shows that AF performs better than the single-task adapter on RTE until removing AF from the first five layers. This improves the inference efficiency by 26\%.\footnote{We include detailed measurements in Appendix~\ref{sec:appendix:adapterdrop-fusion}.}
On CoLA, we observe a different trend. Removing AF from the first layer results in more noticeable performance decreases, achieving lower task performances than the single-task adapter.
This is in line with recent work showing that some linguistic tasks heavily rely on information from the first layers~\citep{vulic-etal-2020-emnlp-probing}. %
We deliberately highlight that AdapterDrop might not be suitable for all tasks. %
However, Figure~\ref{fig:adapter-drop-large} shows that CoLA represents the most extreme case. %
Nevertheless, our results suggest that researchers need to be cautious when removing AdapterFusion layers as there may exist a considerable performance/efficiency tradeoff.

\subsection{AdapterFusion Pruning}

\begin{figure}
	\centering
	\includegraphics[width=0.94\columnwidth]{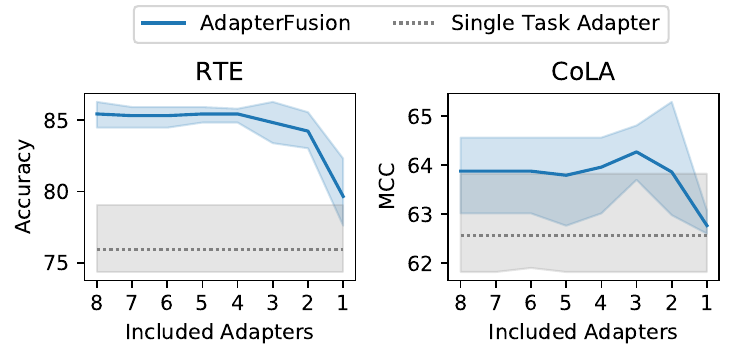}
	\caption{Task performance of AdapterFusion Pruning. AF is trained with eight adapters, and we gradually remove the least important from the model.}
	\label{fig:fusion-pruning}
\end{figure}

The inference efficiency of AF largely depends on the number of fused adapters, see Table~\ref{tab:fusion-speed-adapters}. We can, therefore, achieve efficiency improvements by pruning adapters from the trained AF models (depicted in Figure~\ref{fig:AdapterFusionDrop}). 
Our hypothesis is that we can safely remove adapters if they are not usually activated by AF, which means that they do not contribute much to the output representations. 
In each fusion layer, we record the average adapter activations---their relative importance---using all instances of the respective AF training set. We then remove the adapters with lowest activations. %

Figure~\ref{fig:fusion-pruning} demonstrates that we can remove most adapters in AF without affecting the task performance. With two remaining adapters, we achieve comparable results to the full AF models with eight adapters
and improve the inference speed %
by 68\%.

We therefore recommend performing AdaperFusion pruning before deploying these models in practice. This is a simple yet effective technique to achieve efficiency gains even when aiming at maintaining performance entirely.

\section{Conclusion}
Adapters have emerged as a suitable alternative to full model fine-tuning, and their most widely claimed computational advantage is the small model size. In this work, we have demonstrated that the advantages of adapters go far beyond mere parameter efficiency. Even without our extensions, the training steps of two common adapter architectures are up to 60\% faster. However, these improvements come at the cost of 4--6\% slower inference speed. Thus, if training is more important, adapters can be advantageous over full model fine-tuning.

\emph{AdapterDrop} expands these advantages by dropping a variable number of adapters from lower transformer layers. We \emph{dynamically} 
reduce the computational overhead at run-time when performing inference over multiple tasks and maintain task performances to a large extent. This benefits researchers working on models that need to make multiple independent predictions on a single input.

Finally, we also investigated the computational efficiency of AdapterFusion models. %
We find that dropping entire AdapterFusion layers comes at a considerable performance/efficiency tradeoff, whereas pruning of the least activated adapters in each layer can improve the model efficiency while maintaining performance entirely. 

We believe that our work can be widely extended and that there exist many more directions to obtain efficient adapter-based models. 
For instance, we could explore more efficient pre-trained adapters,\footnote{In Appendix~\ref{app:ssec:adapter_init}, we evaluate MLM pre-trained adapters. Our results suggest that different strategies are necessary for adapters as compared to fully fine-tuned transformers, which can serve as a starting point for further experiments.} sharing the adapter weights across layers,\footnote{Appendix~\ref{appendix:shared-adapter} shows that adapter with shared weights across layers achieves comparable results to a standard adapter while drastically reducing the number of  parameters.} or pruning adapters from AdapterFusion at training time.\footnote{Appendix~\ref{appendix:fusion-training-dropout} shows that we can randomly dropout 75\% of the adapters during AdapterFusion training with a minimal impact on the task performance.}
In the Appendix to this paper, we present preliminary results for several related ideas, which may serve as a starting point for future work.

 \section*{Acknowledgments}
  This work has received financial support from multiple sources. (1) The German Federal Ministry of Education and Research and the Hessian Ministry of Higher Education, Research, Science and the Arts within their joint support of the National Research Center for Applied Cybersecurity ATHENE.
  (2) The European Regional Development Fund (ERDF) and the Hessian State Chancellery – Hessian Minister of Digital Strategy and Development under the promotional reference 20005482 (TexPrax).
  (3) The German Research Foundation (DFG) as part of the Research Training Group KRITIS No. GRK 2222.
  (4) The German Federal Ministry of Education and Research (BMBF) as part of the Software Campus program under the promotional reference 01|S17050.
  (5) The LOEWE initiative (Hesse, Germany) within the emergenCITY center. 
  (6) The German Research Foundation (DFG) as part of the UKP-SQuARE project (grant GU 798/29-1).
 Finally, we gratefully acknowledge the support of NVIDIA Corporation with the donation of the Titan X Pascal GPU used for this research.

\bibliography{anthology,paper}
\bibliographystyle{acl_natbib}

\appendix

\section{Measuring Computational and Task Performance}

\subsection{Computational Efficiency}
\label{sec:appendix:measurements}
We use Python 3.6, PyTorch 1.5.1, CUDA 10.1 for all measurements.
We repeat them with two different GPUs: NVIDIA Tesla V100 PCIe (32GB) and a NVIDIA Titan X Pascal (12GB).   
We make use of the \texttt{torch.cuda.Event} class and \texttt{torch.cuda.synchronize} to measure \textit{only} the exact period of time of a training (or inference) step.\footnote{This is necessary due to the asynchronous nature of the command execution on CPU and GPU.}
For both inference and training, we repeat the respective step 300 times. 
We report the median %
to mitigate the impact of
outliers caused by GPU warmup. %

\paragraph{Relativ speed.}
We define the relative speed of an adapter compared full model fine-tuning as:
$
    \frac{S_a}{S_f}
$
where $S_a$ and $S_f$ are the time of one step with the adapter model and the fully fine-tuned model, respectively.
For example, a relative speed of 1.5 means that the adapter model can perform 1.5 steps in the time the fully fine-tuned model performs one step.

\paragraph{Speedup.}
Speedup describes the positive change in relative speed of an adapter model when using AdapterDrop (or another method). %
A speedup of $p$\% means that the adapter model with AdapterDrop requires only $(1-p/100)\times$ of the runtime than the adapter model without AdapterDrop. %

The speedup
of AdapterDrop (and AdapterFusion) are additive. If dropping one layer results in $p$\% speedup, dropping two layers results in $2p$\% speedup, etc.

\subsection{Task Performances}
\label{sec:appendix:task-performances}
We study the task performances of adapter models on the popular GLUE benchmark~\cite{wang18glue}.
Following \citet{devlin-bert}, we exclude the WNLI because of the problematic data construction.\footnote{See \url{https://gluebenchmark.com/faq}} %
We perform our analyses using RoBERTa base~\citep{liu2019roberta} as our pre-trained model and %
report the mean and standard deviation over three runs of the best development performance evaluated after every epoch. We train larger data sets (SST-2, MNLI, QNLI, and QQP) for 10 epochs and the rest of the data sets for 20 epochs. 
We use a batch size of 32 and, if not otherwise noted, the default hyperparameters for adapter fine-tuning as in~\cite{Pfeiffer2020adapterfusion}.

\section{Adapter Initialization and Convergence}
\label{app:ssec:adapter_init}

Besides measuring training and inference time, we are interested in (1) how using adapters compare to standard RoBERTa-base with regards to downstream task convergence, and (2) if initializing adapters with pre-trained weights using \textit{masked language modeling} can lead to faster convergence.

\begin{figure*}
	\centering
	\includegraphics[width=\linewidth, scale=1.2]{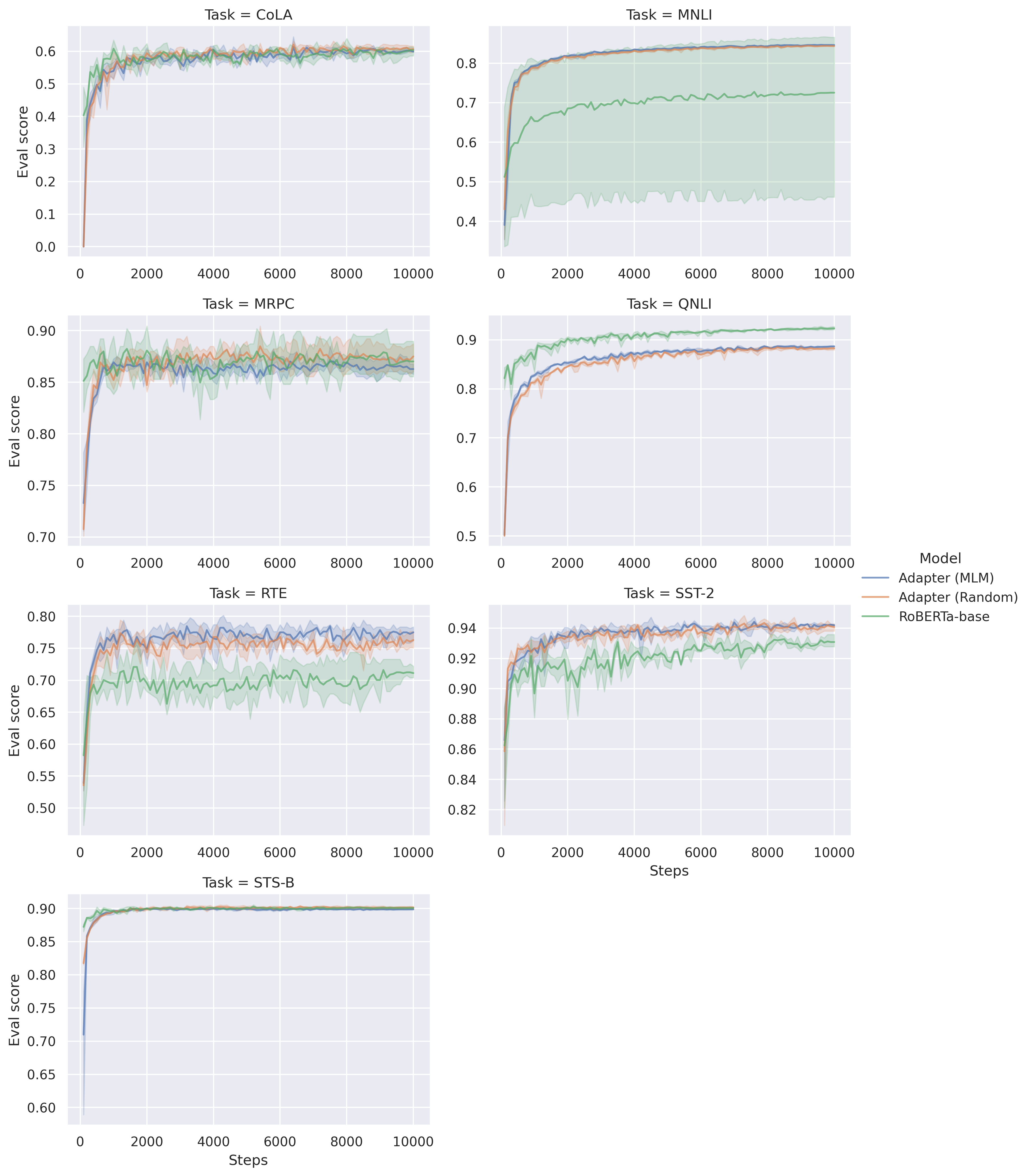}
	\caption{Evaluation performance of fine-tuning RoBERTA-base in comparison with different initialization strategies for adapters (randomly initialized vs. pre-trained on masked language modeling task). Training was conducted for 10k steps with a learning rate of 5e-05 for RoBERTa-base and 0.0001 for adapters, respectively.}
	\label{fig:adapter_init_mlm_performance_diff}
\end{figure*}

First, we compare RoBERTa-base with adapter models using the architecture proposed by \citet{Pfeiffer2020adapterfusion}.
Second, we pretrain an adapter with masked language modeling (MLM) using documents from the English Wikipedia.\footnote{We used a recent dump of English Wikipedia. We train with a batch size of 64 and for 250k steps such that no sentence was used twice.}
The results for both experiments are visualized in Figure~\ref{fig:adapter_init_mlm_performance_diff}.
When comparing RoBERTa-base with randomly initialized adapters, We find that adapters do not come at the cost of requiring more training steps for convergence (1).
For several of the eight GLUE tasks, we observe similar convergence behavior with the standard RoBERTa-base model and its counterpart using adapters.

Further, we observe across all tasks that initializing the adapter weights with MLM pre-training does not have a substantial impact on the downstream task convergence (compared to a randomly initialized adapter). 
Thus, we find no evidence that pre-training of adapters with our masked language modeling objective leads to better convergence performance in our experiments (2).

\section{Detailed Results: AdapterDrop Task Performances}

\begin{figure*}
	\centering
	\includegraphics[width=0.7\textwidth]{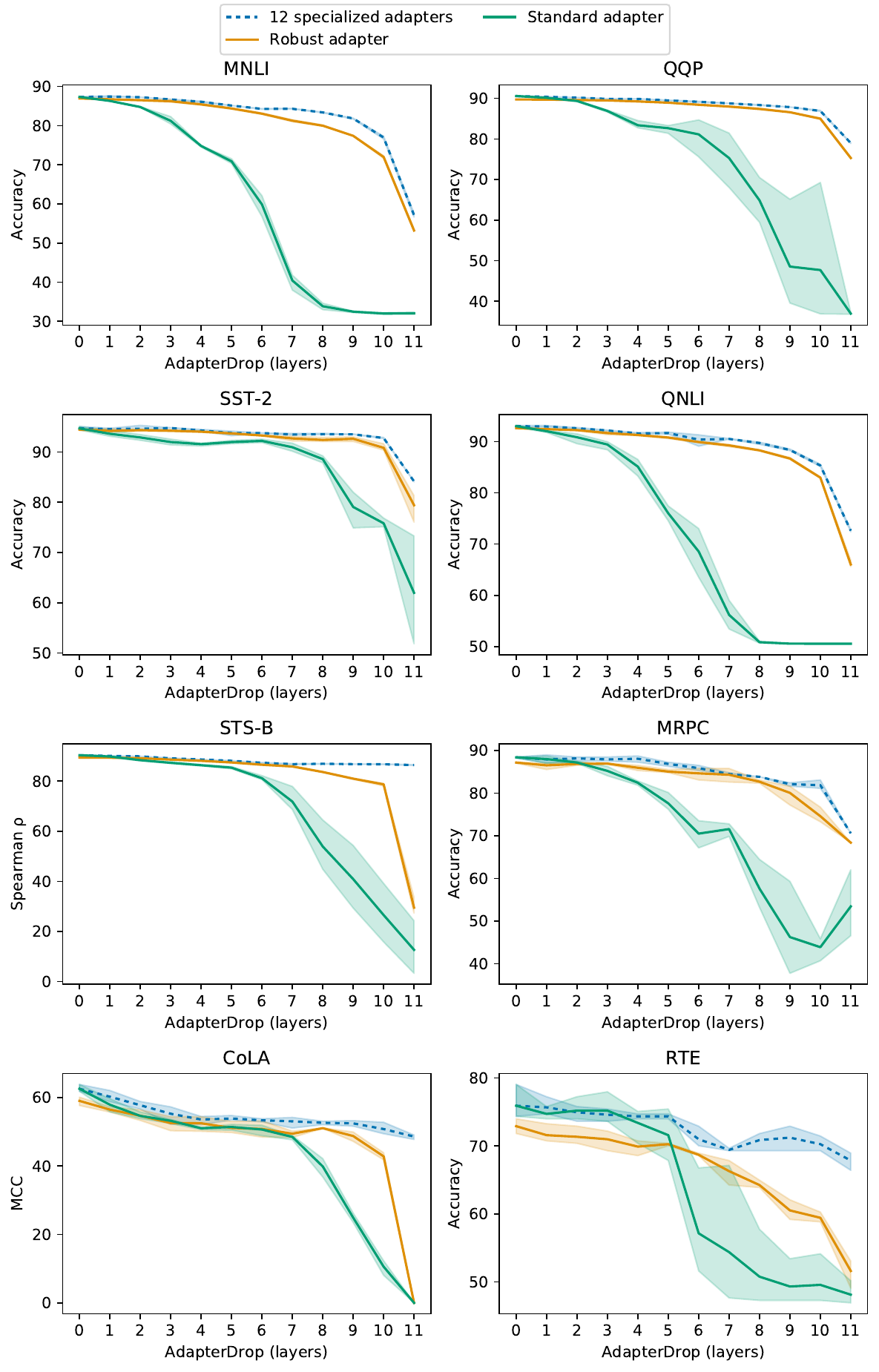}
	\caption{The \textbf{AdapterDrop} task performances for all eight GLUE tasks in relation to the dropped layers. `12 specialized adapters' refers to the performance of indiviudal models trained for each AdapterDrop setting separately (i.e., 12 models); `Standard adapter' refers to the adapter that is trained with no dropped layers; AdapterDrop training refers to the adapter that is trained with our proposed training procedure.}
	\label{fig:adapter-drop-large}
\end{figure*}

We plot the detailed task performances of AdapterDrop with the different training strategies in Figure~\ref{fig:adapter-drop-large}. 
The relative differences of AdapterDrop to a standard adapter with no AdapterDrop are given in Table~\ref{tbl:adapter-drop-maintained-performance}.

\section{Adapter with Cross-Layer Parameter Sharing}
\label{appendix:shared-adapter}
We can further reduce the number of parameters required for each task by sharing the weights of the adapters across all transformer layers. 
This is similar to weight sharing in ALBERT~\citep{lan2019albert}, but specialized on adapters and can therefore be applied to a wide range of pre-trained models.

We use the Pfeiffer adapter architecture in our experiments with the same hyperparameters as in Appendix~\ref{sec:appendix:task-performances}. Because cross-layer parameter sharing reduces the capacity of adapter models, we study the impact of the adapter compression rate. The compression rate refers to the down-projection factor in the adapter’s bottleneck layer and thus impacts the its capacity (the compression rate specifies by how much `FF Down' in Figure~\ref{fig:AdapterDrop} compresses the representations). The standard compression rate is 16, and smaller values result in a larger model capacity.

Table~\ref{tbl:adapter-sharing-all} shows that cross-layer parameter sharing with the same compression rate of 16 largely maintains the performance compared to separate weights with an average difference of 2.35\%. With a smaller compression rate of 4, we close this gap by more than 50\% while still requiring 66\% fewer parameters.\footnote{Even smaller compression rates do not yield similar gains.} %
The resulting models are light-weight: our shared adapter with a compression rate of 16 requires only 307KB storage space.

\section{Training AdapterFusion with Dropout}
\label{appendix:fusion-training-dropout}

We investigate the random dropout of adapters from AdapterFusion during training (using our eight task adapters as in \S\ref{sec:adapter-fusion}) to improve the speed of training steps. Each layer randomly selects different adapters to drop out. This means that the model itself may still use the knowledge from all tasks, although not in the layers individually.

Table~\ref{tbl:appendix:fusion-dropout} shows the results for the four smallest GLUE tasks in terms of training data size. The speedup that we achieve with AdapterFusion dropout can be substantial: with a dropout rate of 75\% (i.e., dropping out 6 out of our 8 adapters) each training step is 74\% faster on average (with a sequence length of 128, a batch size of 32). We observe no clear trend in terms of task performances. Fusion dropout leads to consistent decreases on RTE and CoLA, only a small impact on STS-B (no difference when dropping out 25\% of adapters), and yields improvements on MRPC.

The effectiveness of Fusion dropout, thus, depends on the individual downstream task. Nevertheless, we believe that this methods could be suitable, e.g., for resource-constrained settings.

\section{Detailed Results: Removing AdapterFusion Layers}
\label{sec:appendix-fusion-layer-dropout}
\begin{figure}
	\centering
	\includegraphics[width=\columnwidth]{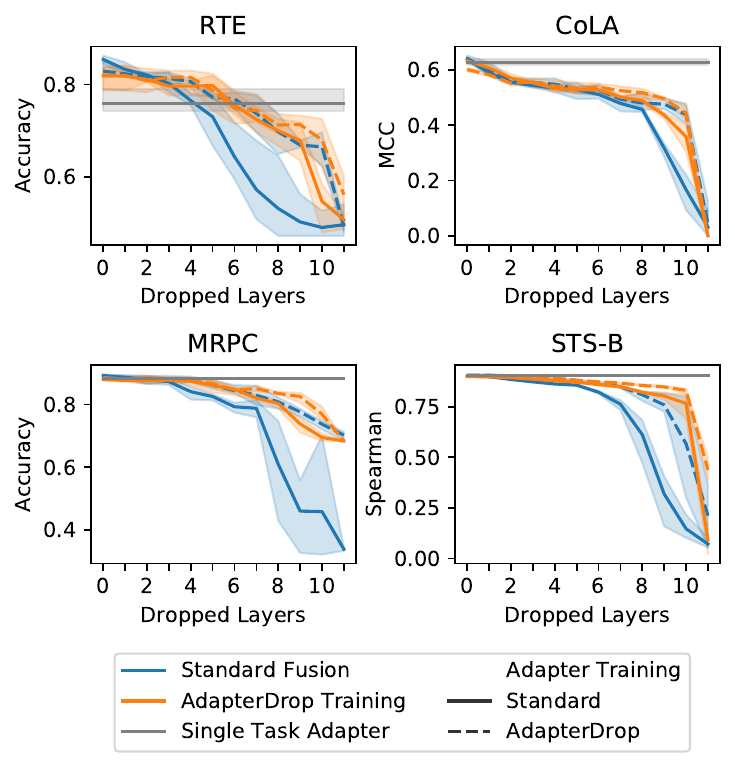}
	\caption{Performance of AF by the number of dropped AF layers. We show the results for AF and the used adapters (both with and without AdapterDrop), and compare the performance with a standard single task adapter.}
	\label{fig:fusion-layer-drop}
\end{figure}

The computational overhead of AF can be reduced during inference by decreasing the number of adapters. We investigate how dropping AF layers impacts the performance on the four smallest GLUE tasks (MRPC, STS-B, CoLA, RTE) and visualize the results in Figure~\ref{fig:fusion-layer-drop}.

In this experiment we compare the performance of AF with and without AdapterDrop during training. For both, we use standard adapters as well as adapters created via AdapterDrop as basis for AF. Unsurprisingly, the performance of AF without AdapterDrop within the adapters or fusion drops fastest on all four datasets. Using AdapterDrop when creating the adapters, applying AdapterDrop on AF, or the combination of both significantly reduces the performance drop when omitting fusion layers during inference. On RTE and MRPC, multiple AF layers can be omitted while still performing en par with or better compared to a single task adapter. We further find this robustness to be task dependent. Even AF with AdapterDrop shows a steep fall in performance on RTE and CoLA, while being relatively stable on MRPC and STS-B, even with most layers omitted. 

\section{Detailed Efficiency Measurements}
In this section, we present detailed results of our efficiency measurements for V100 and TitanX GPUs.

\subsection{Adapters}
\label{sec:appendix:efficiency-adapters}

\begin{figure*}
	\centering
	\subfloat[V100 Inference]{\includegraphics[width=\columnwidth,valign=t]{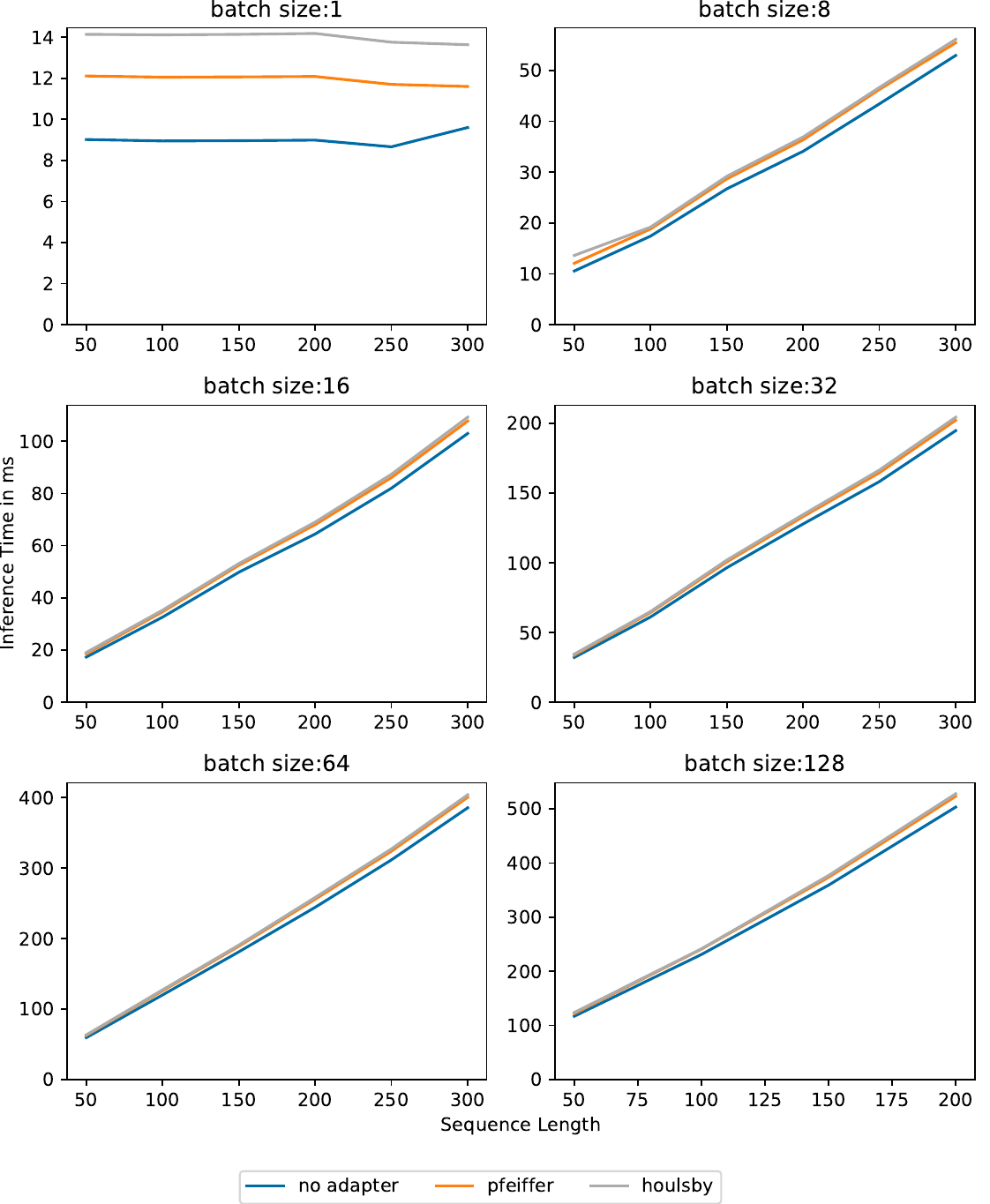}}
	\quad
	\subfloat[TitanX Inference]{\includegraphics[width=\columnwidth,valign=t]{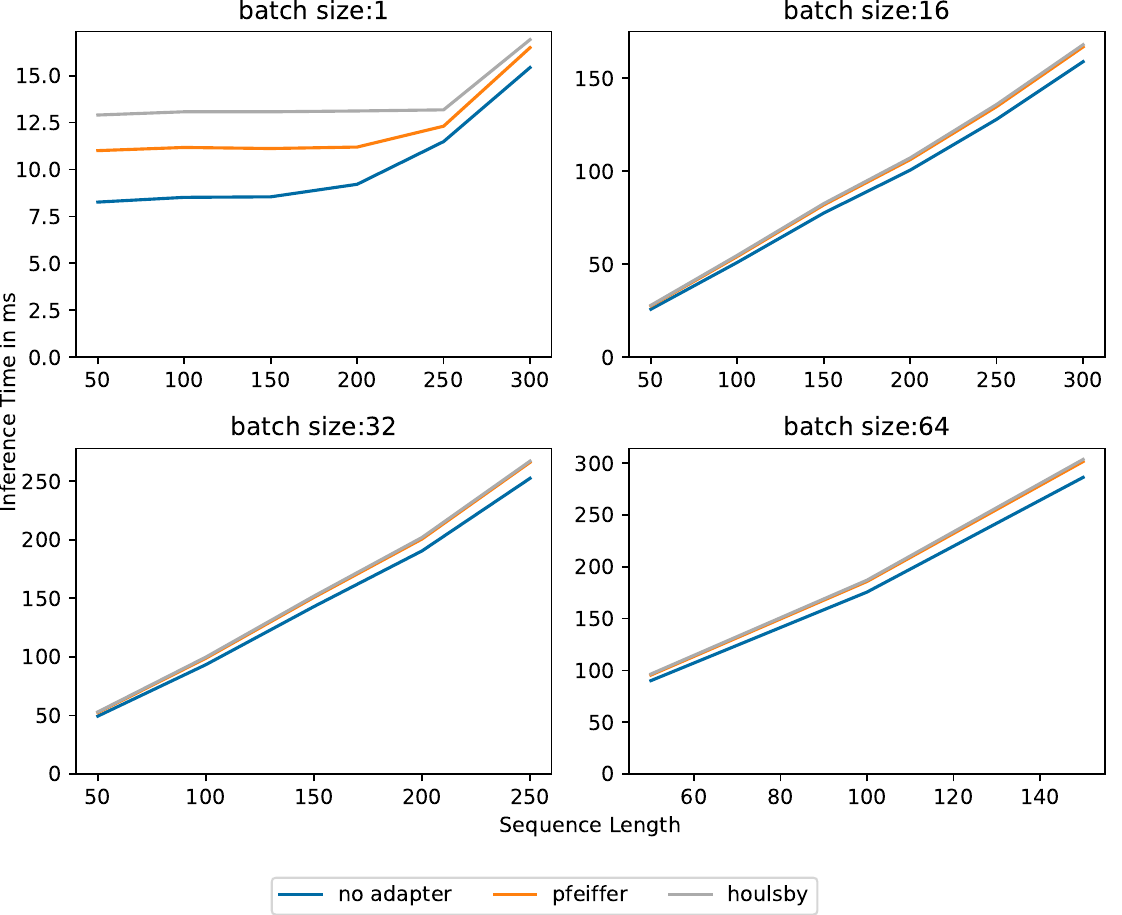}}
	\\
	\subfloat[V100 Training]{\includegraphics[width=\columnwidth,valign=t]{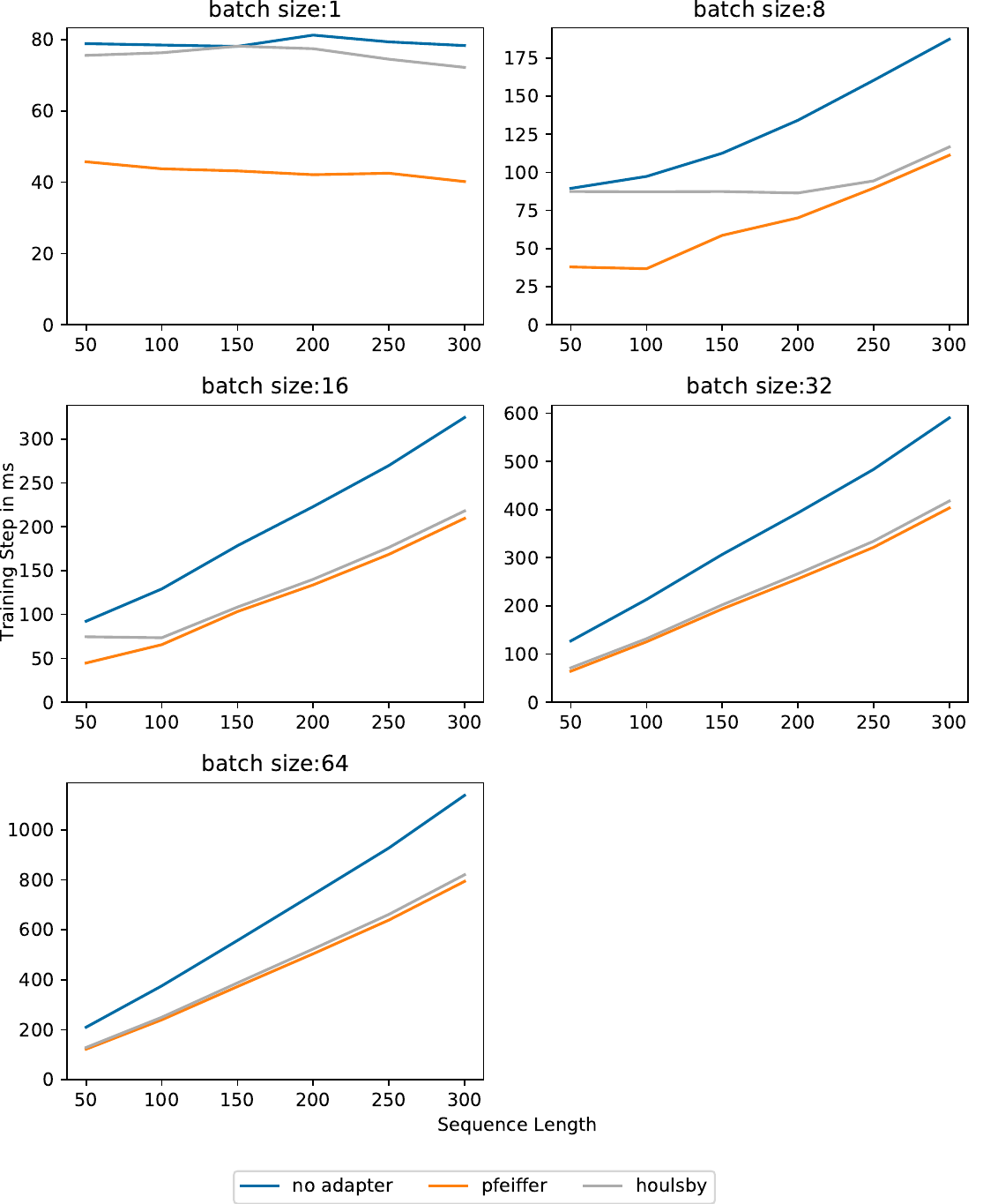}}
	\quad
	\subfloat[TitanX Training]{\includegraphics[width=\columnwidth,valign=t]{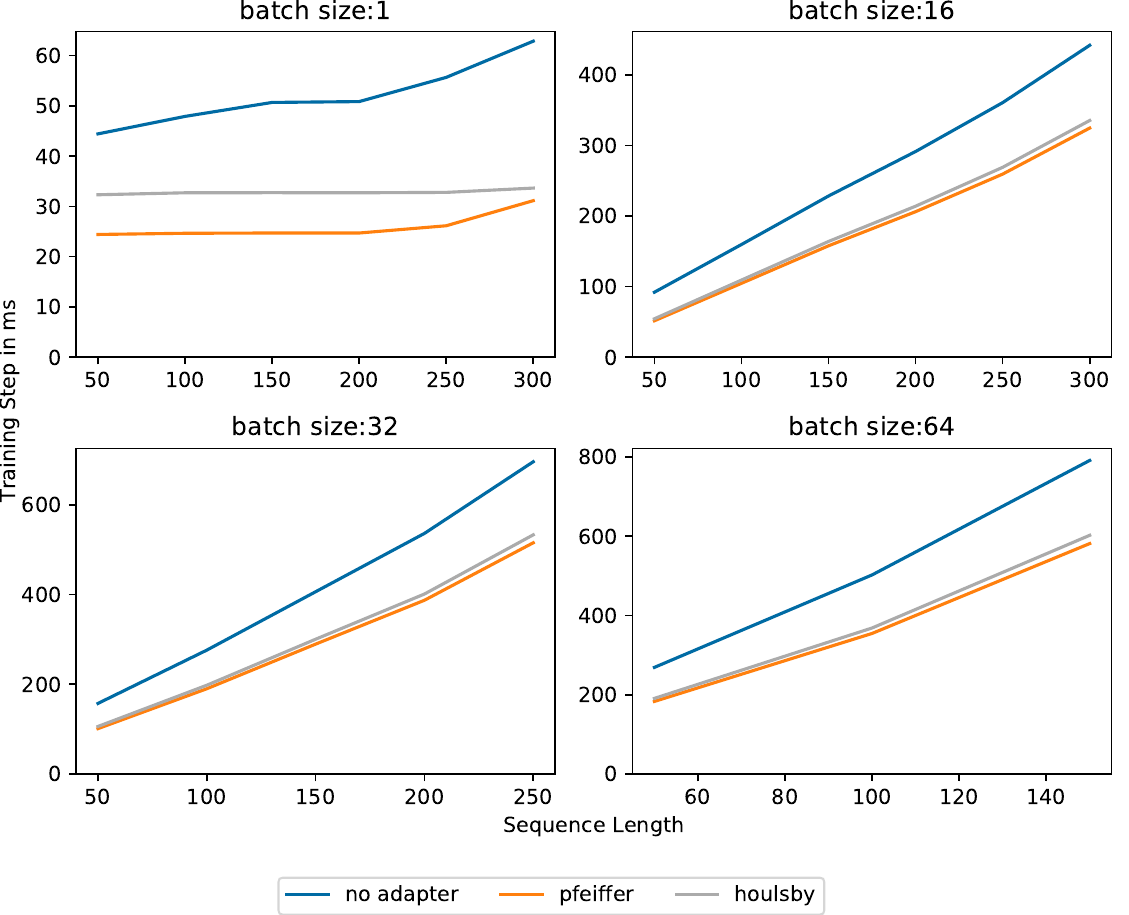}}
	\caption{The absolute time for each \textbf{inference} or \textbf{training step}. We compare a \textbf{transformer model without adapters} and an \textbf{adapter model with Pfeiffer or Houlsby architectures}. We note that for small inputs, i.e., batch size 1 or 8, the time does not increase with the sequence length because the GPU is not working at capacity. Figure (b) with batch size 1 shows the transition from working under and working at capacity.}
	\label{fig:adapter-time-grid}
\end{figure*}

We present the efficiency results for adapters and fully fine-tuned models in Figure \ref{fig:adapter-time-grid}, where we plot the required time (absolute numbers) during training and inference.
The relative speed of adapters compared to fully fine-tuned models is given in Table \ref{tab:adapters:speedup}.

\subsection{AdapterDrop}
\label{sec:appendix:adapterdrop}
\begin{figure*}
	\centering
	\includegraphics[width=1.5\columnwidth]{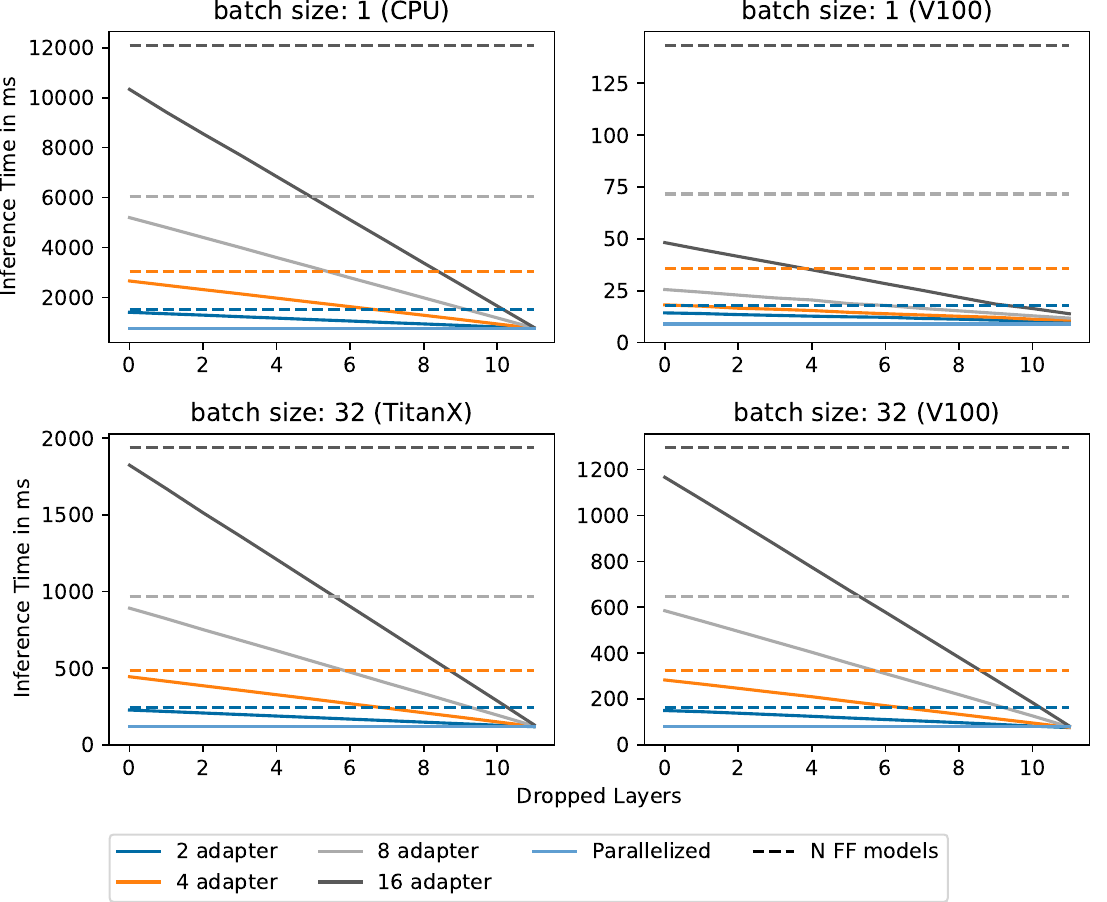}
	\caption{The absolute time required for performing inference for \textbf{multiple tasks} on the same input. %
	The measurements are conducted with a sequence length of 128. \textbf{N FF models} denotes $N$ fully fine-tuned models, executed sequentially. \textbf{Parallelized} denotes the time required by N fully fine-tuned models running fully parallelized. Batch size 1 on the V100 is an outlier compared to the other results with a smaller speedup for each dropped layer but a higher relative speed compared to the fine-tuned models due to the small input size.}
	\label{fig:inference-multiadapter}
\end{figure*}

\paragraph{Multi-task inference.}
In Figure \ref{fig:inference-multiadapter}, we plot the speed of adapters in a multi-task setting compared to fully fine-tuned models with sequential processing of inputs.
In Table \ref{tab:multitask}, we present the relative speed of adapters in this setting and show the speedup gained with AdapterDrop for each dropped layer. 
The average speedup in Table \ref{tab:adapterdrop-speedup} is calculated as the average speedup over the batch sizes 16, 32 and 64 in Table \ref{tab:multitask}. 

\paragraph{Training adapters with dropped layers.}
Table \ref{tab:adapterdrop-training-speedup} shows the speedup of AdapterDrop when training a \emph{single} adapter.
The average speedup for training with AdapterDrop is 4.7\% per layer for the V100 and 4.5\% for the TitanX. This is the average result over batch sizes 16, 32, 64 and sequence length 64, 128, 256, and 256 (see Table \ref{tab:adapterdrop-training-speedup}).

\subsection{AdapterFusion}
\label{sec:appendix:efficiency-adapterfusion}
\begin{figure*}
	\centering
	\subfloat[V100]{\includegraphics[width=1.5\columnwidth]{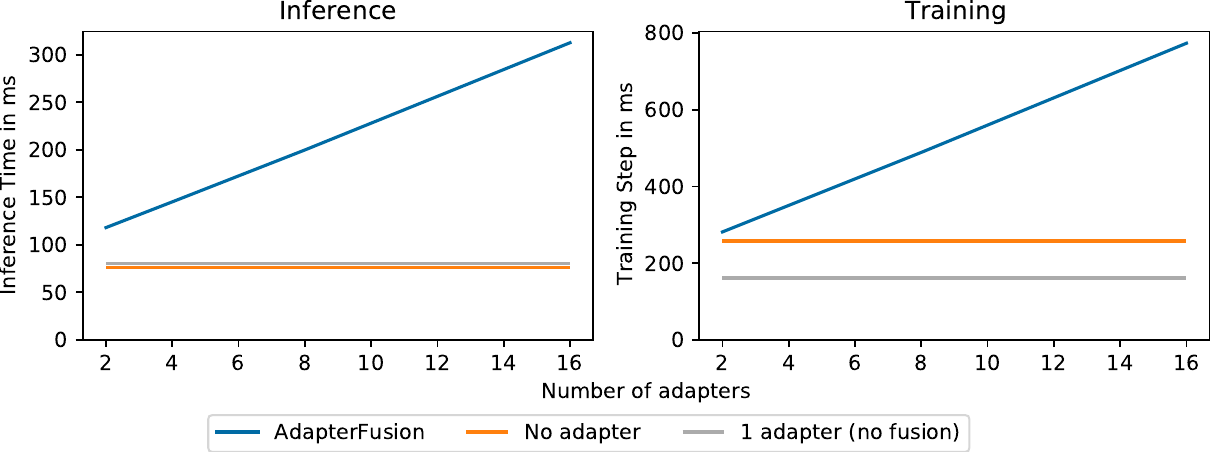}}\\
    \subfloat[TitanX]{\includegraphics[width=1.5\columnwidth]{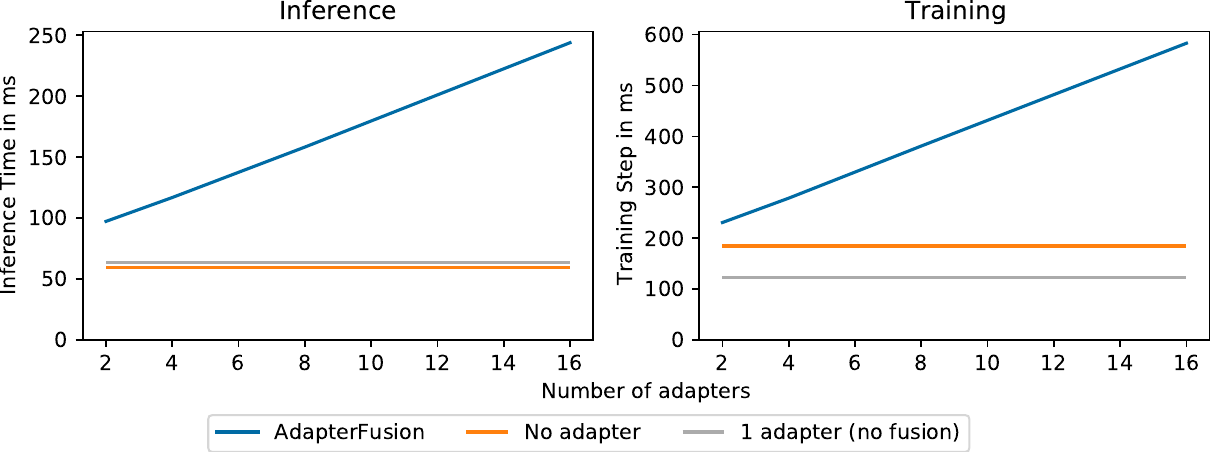}}
	\caption{Absolute time measurements for \textbf{AdapterFusion} at \textbf{inference} (left) and \textbf{training} (right) as a function of the number of adapters. The measurements were conducted with a batch size of 32 (V100) and 16 (TitanX), and a sequence length of 128.}
	\label{fig:fusion:train-inference}
\end{figure*}

We plot the speed of AdapterFusion with different numbers of included adapters in Figure \ref{fig:fusion:train-inference}.
In Table \ref{tab:fusion-speed}, we present the relative speed of AdapterFusion compared to a fully-finetuned model and a model with one adapter. This also shows the computational overhead (slowdown) that results from adding more adapters to AdapterFusion.

\subsection{AdapterDrop for AdapterFusion}
\label{sec:appendix:adapterdrop-fusion}

\begin{figure*}
	\centering
	\subfloat[V100]{\includegraphics[width=1.5\columnwidth]{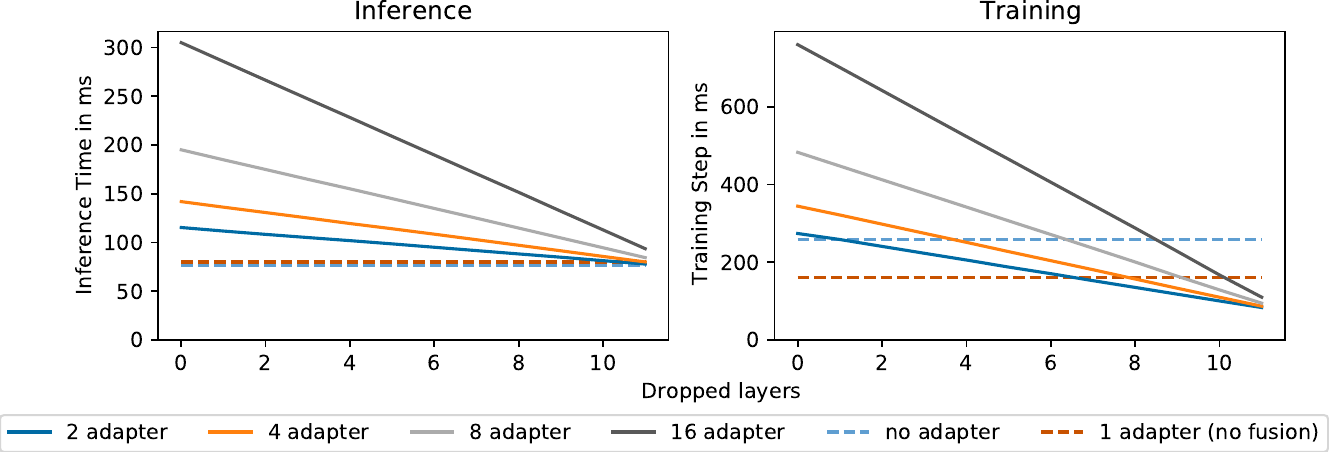}}\\
    \subfloat[TitanX]{\includegraphics[width=1.5\columnwidth]{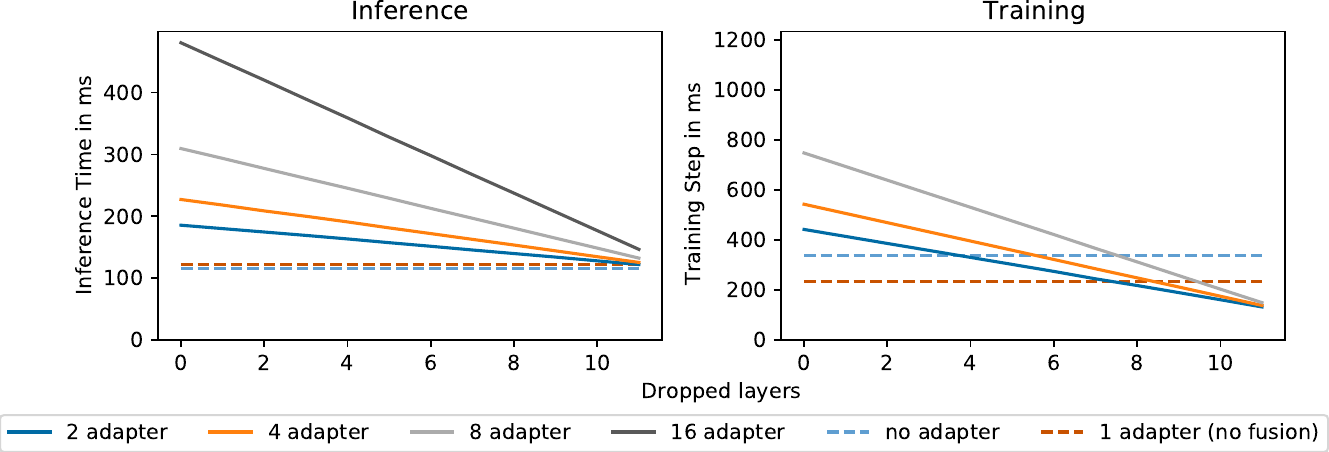}}
	\caption{Absolute time measurements for \textbf{AdapterFusion with AdapterDrop} at \textbf{inference} (left) and \textbf{training} (right) as a function of the number of dropped layers. The measurements were conducted with a batch size of 32 and a sequence length of 128. We additionally plot the time of an adapter (without AdapterDrop) and a model without adapters to provide a more thorough comparison.}
	\label{fig:fusion:layerdrop}
\end{figure*}

Table \ref{tab:adapterdrop-fusion-speedup} shows the speedup gained with AdapterDrop for AdapterFusion during training and inference.
Figure \ref{fig:fusion:layerdrop} shows the required time as a function of the dropped layers.

\section{Parallel Implementation of AdapterFusion}
\label{sec:appendix:parallel-fusion}

AdapterHub's implementation of AdapterFusion passes through each task adapter sequentially.
We hypothesized that a better efficiency can be achieved with parallel processing of adapters.
We implement the parallel computation of the different adapters by reformulation the linear layers as two convolutions. 

The first convolution is a convolution with a kernel size equal to the hidden dimension of the transformer and output channels equal to the number of adapters times the downprojection dimension of the adapters.
The second convolution is a grouped convolution\footnote{Using the 'groups' parameter in Pytorch (\url{https://pytorch.org/docs/stable/generated/torch.nn.Conv1d.html\#torch.nn.Conv1d})} which processes the channels in blocks the size of the downprojection dimension. 
It outputs channels equal to the number of adapters times the hidden dimension.

\begin{figure*}[]
	\centering
	\subfloat[V100]{\includegraphics[width=1.5\columnwidth]{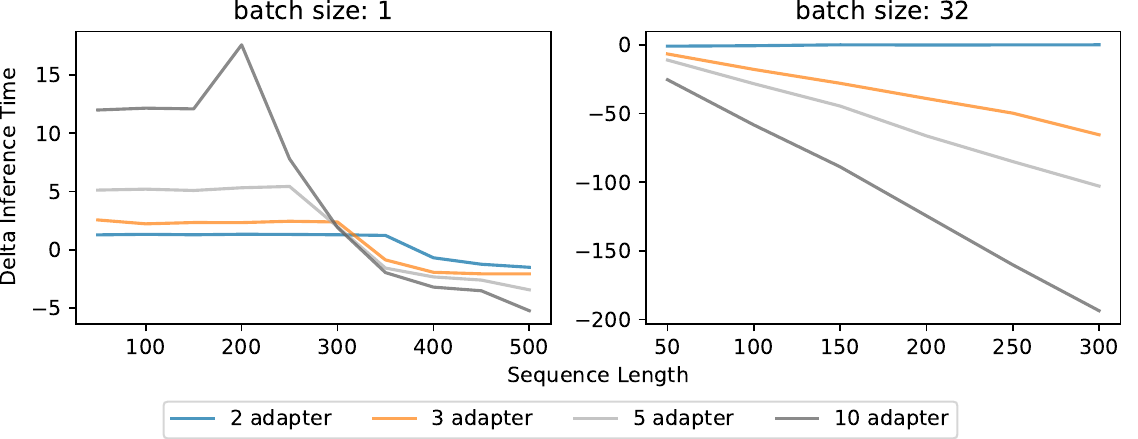}}
	\\
	\subfloat[TitanX]{\includegraphics[width=1.5\columnwidth]{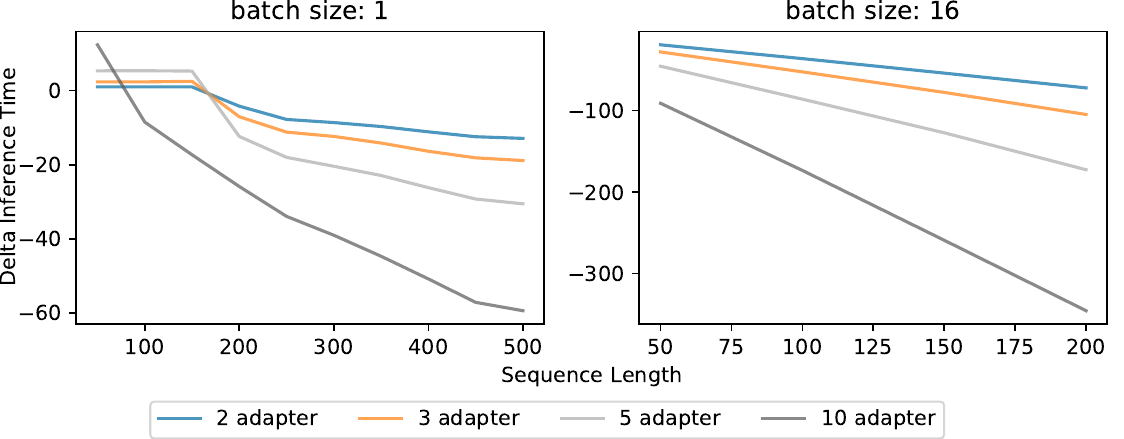}}
	
	\caption{The difference in inference time between \textbf{iterative} and \textbf{parallel} implementations of \textbf{AdapterFusion}. Negative values indicate that the iterative implementation is faster. We calculate the difference as $t_i-t_p$, where $t_i, t_p$ are the times for iterative and parallel implementation, respectively. In Figure (a), the parallel implementation is faster if the input is sufficiently small as the GPU is not working at capacity and is able to use the parallel implementation.}
	\label{fig:delta_iterative_parallel_inference}
\end{figure*}

We show in Figure \ref{fig:delta_iterative_parallel_inference} and in Table \ref{tab:parallel-iterative} that the iterative implementation is \emph{faster} than the parallel implementation for larger input sizes (e.g., batch sizes greater than).
This indicates that once the input can no longer be processed entirely in parallel on the GPU (due to limited CUDA cores) the iterative implementation seems to be more efficient.

\begin{table}
\footnotesize
    \centering
\begin{tabular}{rrrrr}
\toprule
  & \multicolumn{4}{c}{\textbf{Speedup} (per dropped layer)} \\
  \cmidrule(lr){2-5}
  & \multicolumn{2}{c}{\bf Inference} & \multicolumn{2}{c}{\bf Training} \\
  \cmidrule(lr){2-3} \cmidrule(lr){4-5}
\bf Adapters &  \bf V100 & \bf TitanX &  \bf V100 & \bf TitanX\\
\midrule
            2 & 3.0\% & 3.1\% & 6.3\% & 6.4\% \\
            4 & 4.0\% & 4.1\% & 6.8\% & 6.8\% \\
            8 & 5.2\% & 5.2\% & 7.3\% & 7.3\% \\
           16 & 6.3\% & 6.3\% & 7.8\% &    - \\
\bottomrule
\end{tabular}
    \caption{The speedup for each dropped layer for \textbf{AdapterFusion} during training and inference. Measurements were conducted with a batch size of 32 and sequence length of 128. Missing values are due to insufficient GPU memory.
    }
    \label{tab:adapterdrop-fusion-speedup}
\end{table}

\begin{table}
\footnotesize
    \centering
\begin{tabular}{rrrr}
\toprule
  & & \multicolumn{2}{c}{\bf Speedup} \\
  \cmidrule(lr){3-4}
\bf Batch Size & \bf  Seq. Len &  \bf V100 & \bf TitanX\\
\midrule
    16 &   64 & 4.6\% &   4.4\% \\
    16 &  128 & 4.6\% &   4.6\% \\
    16 &  256 & 4.8\% &   4.6\% \\
    16 &  512 & 4.7\% &      - \\
    32 &   64 & 4.6\% &   4.5\% \\
    32 &  128 & 4.7\% &   4.5\% \\
    32 &  256 & 4.6\% &   4.7\% \\
    32 &  512 & 4.8\% &      - \\
    64 &   64 & 4.7\% &   4.5\% \\
    64 &  128 & 4.6\% &   4.5\% \\
    64 &  256 & 4.7\% &      - \\
    64 &  512 &    - &      - \\
\bottomrule
\end{tabular}
    \caption{Speedup for each dropped layer during \textbf{training with AdapterDrop} on the V100 and TitanX.
    }
    \label{tab:adapterdrop-training-speedup}
\end{table}

\begin{table}
    \centering
    \footnotesize
    \setlength{\tabcolsep}{4pt}
    \begin{tabular}{lrrrr}
    \toprule
     & \bf Standard & \multicolumn{3}{c}{\bf Cross-Layer Parameter Sharing} \\ 
    \cmidrule(lr){2-2} \cmidrule(lr){3-5}
    \multicolumn{2}{r}{Compression~rate\,=\,16} & 1.33 & 4 & 16 \\
    \midrule 
    \bf SST-2 & $94.7$ \stdintable{0.3} & $94.2$ \stdintable{0.3} & $94.2$ \stdintable{0.1} & $94.1$ \stdintable{0.4} \\
    \bf QNLI & $93.0$ \stdintable{0.2} & $92.4$ \stdintable{0.1} & $93.1$ \stdintable{0.1} & $90.6$ \stdintable{1.4} \\
    \bf MNLI & $87.3$ \stdintable{0.1} & $87.0$ \stdintable{0.1} & $87.1$ \stdintable{0.0} & $86.2$ \stdintable{0.2} \\
    \bf QQP & $90.6$ \stdintable{0.0} & 90.8 \stdintable{0.1}  & $90.2$ \stdintable{0.0} & $88.6$ \stdintable{0.5} \\
    \bf CoLA & $62.6$ \stdintable{0.9} & $60.3$ \stdintable{1.6} & $60.8$ \stdintable{0.4} & $57.2$ \stdintable{1.0} \\
    \bf MRPC & $88.4$ \stdintable{0.1} & $88.2$ \stdintable{0.7} & $88.5$ \stdintable{1.1} & $86.8$ \stdintable{0.5} \\
    \bf RTE & $75.9$ \stdintable{2.2} & $69.4$ \stdintable{0.5} & $71.5$ \stdintable{2.7} & $71.5$ \stdintable{1.0}  \\
    \bf STS-B & $90.3$ \stdintable{0.1} & $89.5$ \stdintable{0.1} & $89.7$ \stdintable{0.3} & $89.0$ \stdintable{0.7} \\
    \midrule
    \bf Average & 85.35 & 83.98 & 84.39 & 83.0 \\
    \midrule
    \textit{Params} & \textit{884k} & \textit{884k} & \textit{295k} & \textit{74k} \\
    \bottomrule
    \end{tabular}
    \caption{Task performance scores of the standard approach with separate adapter weights vs. \textbf{cross-layer parameter sharing}. The compression rate denotes the factor by which `FF Down' in Figure~\ref{fig:AdapterDrop} compresses the representations. The number of parameters is given without classification heads.
    }
    \label{tbl:adapter-sharing-all}
\end{table}

\begin{table}
    \centering
    \footnotesize
    \setlength{\tabcolsep}{3.5pt}
    \begin{tabular}{lrrrr}
    \toprule
     & \multicolumn{4}{c}{\bf Fusion Dropout} \\ 
    \cmidrule(lr){2-5}
    & \multicolumn{1}{c}{0\%} & \multicolumn{1}{c}{25\%} & \multicolumn{1}{c}{50\%} & \multicolumn{1}{c}{75\%} \\
    \midrule 
    \bf CoLA & 63.9 \stdintable{0.6} & 62.9 \stdintable{0.8} & 62.4 \stdintable{0.7} &  60.4 \stdintable{0.2} \\
    \bf MRPC & 88.4 \stdintable{0.1} & 89.2 \stdintable{0.5} & 89.2 \stdintable{0.4} &  89.3 \stdintable{0.1} \\
    \bf RTE & 85.4 \stdintable{0.7} & 82.8 \stdintable{1.9} & 82.1 \stdintable{0.3} &  80.9 \stdintable{1.1} \\
    \bf STS-B & 90.2 \stdintable{0.1} & 90.2 \stdintable{0.1} & 90.1 \stdintable{0.1} &  89.9 \stdintable{0.1} \\
    \midrule
    Speedup (8) & - & 15.9\% & 39.4\%  &73.7\%  \\
    Speedup (16) & - & 22.5\%  &  58.2\%  &  120.6\% \\
    \bottomrule
    \end{tabular}
    \caption{Development scores of \textbf{AdapterFusion} (compression rate 16x) with or without fusion dropout during \emph{training}. Fusion dropout of 50\% means that each adapter has a 50\% chance of not being used as input to the fusion layer. The \textbf{speedup} depends on the total number of adapters used in AdapterFusion (8 adapters in our setting here, 16 used by \citet{Pfeiffer2020adapterfusion})} 
    \label{tbl:appendix:fusion-dropout}
\end{table}

\begin{table*}
    \centering
    \footnotesize
    \setlength{\tabcolsep}{4.5pt} 
    \begin{tabular}{lrrrrrrrrrrrr}
    \toprule
     & \multicolumn{12}{c}{\bf Dropped Layers} \\ 
    \cmidrule(lr){2-13}
    & 0 & 1 & 2 & 3 & 4 & 5 & 6 & 7 & 8 & 9 & 10 & 11 \\
    \midrule 
    \bf Standard adapter &
    100.0 & 
    98.5 & 
    97.1 & 
    95.3 & 
    92.0 & 
    89.0 & 
    82.2 & 
    74.6 & 
    64.5 & 
    54.5 & 
    49.3 & 
    43.3 
    \\
    \midrule
    \bf Specialized AdapterDrop (12 models) & 
    100.0 & 
    99.5 & 
    98.9 & 
    98.2 & 
    97.6 & 
    97.1 & 
    95.9 & 
    95.3 & 
    95.1 & 
    94.3 & 
    92.5 & 
    82.9
    \\
    \bf Robust AdapterDrop & 
    98.5 &
    97.7 &
    97.3 &
    96.8 &
    96.1 &
    95.4 &
    94.5 &
    93.3 &
    92.2 &
    89.9 &
    85.9 &
    62.0 
    \\
    \bottomrule
\end{tabular}
\caption{Model performances with \textbf{AdapterDrop} in relation to a standard adapter with no dropped layers. We report the percentage of retained task performance compared to the standard adapter with no dropped layers during evaluation. The results are averaged over all eight GLUE task. A value of 97.1 for specialized AdapterDrop with five dropped layers means that the model achieves 97.1\% of the performance compared to the standard adapter with no dropped layers. Performance scores for each task can be found in Figure~\ref{fig:adapter-drop-large}.} 
\label{tbl:adapter-drop-maintained-performance}
\end{table*}

\begin{table*}
\footnotesize
    \centering
    \setlength{\tabcolsep}{5pt}
\begin{tabular}{rrrrrrrrrr}
\toprule
& & \multicolumn{4}{c}{\bf V100} & \multicolumn{4}{c}{\bf TitanX} \\
\cmidrule(lr){3-6} \cmidrule(lr){7-10}
 \bf Sequence Len. & \bf Batch Size & \multicolumn{2}{c}{\bf Training} & \multicolumn{2}{c}{\bf Inference} & \multicolumn{2}{c}{\bf Training} & \multicolumn{2}{c}{\bf Inference} \\
 \cmidrule(lr){3-4} \cmidrule(lr){5-6} \cmidrule(lr){7-8} \cmidrule(lr){9-10}
  &  & Houlsby & Pfeiffer & Houlsby & Pfeiffer & Houlsby & Pfeiffer & Houlsby & Pfeiffer \\
\midrule
              64 &     16 & 0.98 & 1.70 & 0.92 & 0.94 & 1.61 & 1.69 & 0.93 & 0.94 \\
              64 &     32 & 1.70 & 1.81 & 0.94 & 0.95 & 1.48 & 1.55 & 0.93 & 0.94 \\
              64 &     64 & 1.46 & 1.54 & 0.94 & 0.95 & 1.40 & 1.46 & 0.94 & 0.94 \\
              64 &    128 & 1.48 & 1.55 & 0.95 & 0.96 & 1.37 & 1.42 & 0.94 & 0.94 \\
             128 &     16 & 1.48 & 1.57 & 0.94 & 0.95 & 1.45 & 1.52 & 0.93 & 0.94 \\
             128 &     32 & 1.53 & 1.60 & 0.94 & 0.95 & 1.38 & 1.44 & 0.94 & 0.95 \\
             128 &     64 & 1.47 & 1.53 & 0.95 & 0.96 & 1.35 & 1.40 & 0.94 & 0.95 \\
             128 &    128 & 1.42 & 1.48 & 0.95 & 0.96 &    - &    - &    - &    - \\
             256 &     16 & 1.42 & 1.49 & 0.94 & 0.95 & 1.34 & 1.38 & 0.94 & 0.95 \\
             256 &     32 & 1.40 & 1.46 & 0.95 & 0.96 & 1.31 & 1.36 & 0.94 & 0.96 \\
             256 &     64 & 1.40 & 1.45 & 0.95 & 0.96 &    - &    - &    - &    - \\
             256 &    128 &    - &    - &    - &    - &    - &    - &    - &    - \\
             512 &     16 & 1.36 & 1.41 & 0.96 & 0.96 &    - &    - &    - &    - \\
             512 &     32 & 1.33 & 1.37 & 0.96 & 0.96 &    - &    - &    - &    - \\
             512 &     64 &    - &    - &    - &    - &    - &    - &    - &    - \\
             512 &    128 &    - &    - &    - &    - &    - &    - &    - &    - \\
\bottomrule
\end{tabular}
    \caption{Relative speed of \textbf{adapters} compared to fully fine-tuned models. Missing values are due to insufficient GPU memory.
    }
    \label{tab:adapters:speedup}
\end{table*}

\begin{table*}
\footnotesize
    \centering
\begin{tabular}{rrrrrr >{\columncolor[gray]{0.95}} r >{\columncolor[gray]{0.95}} rrrrr >{\columncolor[gray]{0.95}} r >{\columncolor[gray]{0.95}} r}
\toprule
\rowcolor{white}
& & \multicolumn{6}{c}{\bf V100} & \multicolumn{6}{c}{\bf TitanX} \\
\cmidrule(lr){3-8} \cmidrule(lr){9-14}
\rowcolor{white}
  & & \multicolumn{2}{c}{\bf vs. FF} & \multicolumn{2}{c}{\bf vs. Adap.} & \multicolumn{2}{c}{\bf Slowdown} & \multicolumn{2}{c}{\bf vs. FF} & \multicolumn{2}{c}{\bf vs. Adap} & \multicolumn{2}{c}{\bf Slowdown}\\
  \cmidrule(lr){3-4} \cmidrule(l){5-6} \cmidrule(l){7-8} \cmidrule(lr){9-10} \cmidrule(l){11-12} \cmidrule(l){13-14}
  \rowcolor{white}
\bf Seq. Len & \bf Batch Size & Tr. & Inf. & Tr. & Inf. & Tr. & Inf. & Tr. & Inf. & Tr. & Inf. & Tr. & Inf. \\
\midrule
              64 &     16 & 0.77 & 0.62 & 0.45 & 0.66 &   8.2\% &  10.6\% & 0.88 & 0.62 & 0.52 & 0.66 &  10.3\% &  10.2\% \\
              64 &     32 & 1.03 & 0.64 & 0.57 & 0.68 &  12.0\% &  11.1\% & 0.80 & 0.61 & 0.52 & 0.64 &  11.2\% &  11.0\% \\
              64 &     64 & 0.87 & 0.64 & 0.57 & 0.67 &  12.6\% &  12.0\% & 0.76 & 0.61 & 0.52 & 0.65 &  11.6\% &  11.4\% \\
             128 &     16 & 0.91 & 0.65 & 0.58 & 0.69 &  12.0\% &  11.0\% & 0.80 & 0.61 & 0.53 & 0.65 &  10.9\% &  10.8\% \\
             128 &     32 & 0.92 & 0.64 & 0.57 & 0.68 &  12.5\% &  11.8\% & 0.76 & 0.62 & 0.53 & 0.66 &  11.4\% &  11.1\% \\
             128 &     64 & 0.87 & 0.65 & 0.57 & 0.68 &  12.5\% &  11.6\% &    - &    - &    - &    - &      - &      - \\
             256 &     16 & 0.88 & 0.66 & 0.59 & 0.69 &  12.1\% &  11.3\% & 0.77 & 0.65 & 0.56 & 0.68 &  10.8\% &  10.4\% \\
             256 &     32 & 0.86 & 0.68 & 0.59 & 0.70 &  11.9\% &  11.3\% &    - &    - &    - &    - &      - &      - \\
             256 &     64 &    - &    - &    - &    - &      - &      - &    - &    - &    - &    - &      - &      - \\
             512 &     16 & 0.87 & 0.69 & 0.62 & 0.72 &  11.2\% &  10.1\% &    - &    - &    - &    - &      - &      - \\
             512 &     32 &    - &    - &    - &    - &      - &      - &    - &    - &    - &    - &      - &      - \\
             512 &     64 &    - &    - &    - &    - &      - &      - &    - &    - &    - &    - &      - &      - \\
\bottomrule
\end{tabular}
    \caption{Relative speed of \textbf{AdapterFusion} for different sequence lengths and batch sizes. We compute the training (\textbf{Tr}.) speed and inference (\textbf{Inf}.) speed with two adapters in AdapterFusion. We compare this to: \textbf{FF}, a fully fine-tuned model; \textbf{Adap}, an adapter model (Pfeiffer architecture). The \textbf{slowdown} denotes the computational overhead of \emph{each additional adapter} composed in AdapterFusion (calculated as the average slowdown for adding one adapter to AF consisting of 2--16 adapters).  Missing values are due to insufficient GPU memory.
    }
    \label{tab:fusion-speed}
\end{table*}

\begin{table}
\footnotesize
    \centering
    \setlength{\tabcolsep}{4pt}
\begin{tabular}{lrrr >{\columncolor[gray]{0.95}} r}
\toprule
\rowcolor{white}
\bf Device & \bf Batch Size & \bf Adapters &  \bf Inference & \bf Speedup\\
\midrule
\multirow{16}{*}{\bf V100}
 &       1 &         2 &              1.25 &           2.6\% \\
 &       1 &         4 &              1.97 &           3.7\% \\
 &       1 &         8 &              2.80 &           4.9\% \\
 &       1 &        16 &              2.97 &           6.5\% \\
 &      16 &         2 &              1.13 &           4.1\% \\
 &      16 &         4 &              1.14 &           6.5\% \\
 &      16 &         8 &              1.20 &           7.7\% \\
 &      16 &        16 &              1.16 &           8.4\% \\
 &      32 &         2 &              1.08 &           4.5\% \\
 &      32 &         4 &              1.14 &           6.6\% \\
 &      32 &         8 &              1.11 &           7.9\% \\
 &      32 &        16 &              1.11 &           8.5\% \\
 &      64 &         2 &              1.08 &           4.3\% \\
 &      64 &         4 &              1.05 &           6.7\% \\
 &      64 &         8 &              1.06 &           7.9\% \\
 &      64 &        16 &              1.06 &           8.4\% \\
    \midrule
\multirow{4}{*}{\bf TitanX}
 &          32 &         2 &              1.07 &           4.4\% \\
 &      32 &         4 &              1.09 &           6.6\% \\
 &      32 &         8 &              1.09 &           7.8\% \\
 &      32 &        16 &              1.06 &           8.4\% \\
    \midrule
\multirow{4}{*}{\bf CPU}
 &       1 &             2 &              0.98 &           4.2\% \\
 &       1 &         4 &              1.03 &           6.5\% \\
 &       1 &         8 &              1.05 &           7.7\% \\
 &       1 &        16 &              1.06 &           8.4\% \\
\bottomrule
\end{tabular}
    \caption{The relative inference speed of simultaneous processing of multiple tasks with adapters compared to sequential processing of tasks with fully fine-tuned models. Gray columns show the speedup of \textbf{AdapterDrop} for \emph{every} additional dropped layer. All measurements use a sequence length of 128. Batch size 1 for the V100 is an outlier in both speedup and relative speed compared to the other results due to the small input size (compare with Figure \ref{fig:inference-multiadapter}).%
    }
    \label{tab:multitask}
\end{table}

\begin{table}
\footnotesize
    \centering
\begin{tabular}{rrrrr}
\toprule
&&& \multicolumn{2}{c}{\bf Rel. Speed} \\
\cmidrule{4-5}
\bf Adapters & \bf Seq. Len & \bf Batch Size & \bf V100 & \bf TitanX\\
\midrule
            2 &              100 &      1 & 0.93 & 0.94 \\
            3 &              100 &      1 & 0.89 & 0.88 \\
            5 &              100 &      1 & 0.77 & 0.76 \\
           10 &              100 &      1 & 0.60 & 1.29 \\
            2 &              100 &     16 & 1.02 & 1.44 \\
            3 &              100 &     16 & 1.12 & 1.58 \\
            5 &              100 &     16 & 1.17 & 1.80 \\
           10 &              100 &     16 & 1.27 & 2.14 \\
            2 &              100 &     32 & 1.01 & 1.48 \\
            3 &              100 &     32 & 1.17 & 1.62 \\
            5 &              100 &     32 & 1.23 & 1.85 \\
           10 &              100 &     32 & 1.32 & 2.24 \\
            2 &              200 &      1 & 0.93 & 1.24 \\
            3 &              200 &      1 & 0.88 & 1.37 \\
            5 &              200 &      1 & 0.77 & 1.55 \\
           10 &              200 &      1 & 0.52 & 1.87 \\
            2 &              200 &     16 & 1.01 & 1.46 \\
            3 &              200 &     16 & 1.17 & 1.59 \\
            5 &              200 &     16 & 1.23 & 1.82 \\
           10 &              200 &     16 & 1.32 & 2.21 \\
            2 &              200 &     32 & 1.00 & 1.11 \\
            3 &              200 &     32 & 1.18 & 1.17 \\
            5 &              200 &     32 & 1.26 &    - \\
           10 &              200 &     32 & 1.34 &    - \\
            2 &              300 &      1 & 0.93 & 1.37 \\
            3 &              300 &      1 & 0.88 & 1.50 \\
            5 &              300 &      1 & 0.91 & 1.70 \\
           10 &              300 &      1 & 0.94 & 2.03 \\
            2 &              300 &     16 & 1.00 & 1.48 \\
            3 &              300 &     16 & 1.16 & 1.63 \\
            5 &              300 &     16 & 1.22 & 1.88 \\
           10 &              300 &     16 & 1.32 &    - \\
            2 &              300 &     32 & 1.00 &    - \\
            3 &              300 &     32 & 1.20 &    - \\
            5 &              300 &     32 & 1.27 &    - \\
           10 &              300 &     32 & 1.36 &    - \\
            2 &              400 &      1 & 1.04 & 1.39 \\
            3 &              400 &      1 & 1.09 & 1.51 \\
            5 &              400 &      1 & 1.10 & 1.74 \\
           10 &              400 &      1 & 1.10 & 2.08 \\
            2 &              400 &     16 & 1.00 &    - \\
            3 &              400 &     16 & 1.18 &    - \\
            5 &              400 &     16 & 1.25 &    - \\
           10 &              400 &     16 & 1.34 &    - \\
            2 &              400 &     32 & 1.00 &    - \\
            3 &              400 &     32 & 1.20 &    - \\
            5 &              400 &     32 & 1.27 &    - \\
           10 &              400 &     32 &    - &    - \\
\bottomrule
\end{tabular}
    \caption{Relative speed of \textbf{AdapterFusion} with the \textbf{iterative} implementation versus the \textbf{parallel} implementation with different batch sizes, sequence lengths and numbers of adapters for the V100 and TitanX. The parallel implementation is faster if the input is sufficiently small (batch size 1 or 2 adapters) as the GPU is not working at capacity and is able to use the parallel implementation.
    }
    \label{tab:parallel-iterative}
\end{table}

\end{document}